\documentclass[10pt,twocolumn,letterpaper]{article}

\usepackage{cvpr}      

\usepackage{booktabs}
\usepackage{multirow}
\usepackage{graphicx}      
\usepackage{adjustbox}     

\usepackage{geometry}
\usepackage{amsmath,amssymb}
\usepackage{enumitem}
\usepackage{algorithm}
\usepackage{algpseudocode}

\usepackage[table]{xcolor} 

\definecolor{cvprblue}{rgb}{0.21,0.49,0.74}
\usepackage[pagebackref,breaklinks,colorlinks,allcolors=cvprblue]{hyperref}

\def\paperID{4869} 
\def\confName{CVPR}
\def\confYear{2026}

\title{Generative Human-Object Interaction Detection via Differentiable Cognitive Steering of Multi-modal LLMs}

\author{Zhaolin Cai\textsuperscript{1}, Huiyu Duan\textsuperscript{1}\textsuperscript{†}, Zitong Xu\textsuperscript{1}, Fan Li\textsuperscript{2}, Zhi Liu\textsuperscript{3}, \\Jing Liu\textsuperscript{4}, Wei Shen\textsuperscript{3},Xiongkuo Min\textsuperscript{1}\textsuperscript{†}\, Guangtao Zhai\textsuperscript{1}\textsuperscript{†}\\
\textsuperscript{1}Institute of Image Communication and Network Engineering, Shanghai Jiao Tong University\\
\textsuperscript{2}Xi'an Jiao Tong University, \textsuperscript{3}Shandong University, \textsuperscript{4}Tianjin University\\
$\{$huiyuduan, xuzitong zhangkaiwei, qiang.hu, minxiongkuo, zhaiguangtao$\}$@sjtu.edu.cn
}

\begin{document}
\maketitle
\begin{abstract}
\vspace{-1mm}
Human-object interaction (HOI) detection aims to localize human-object pairs and the interactions between them. Existing methods operate under a closed-world assumption, treating the task as a classification problem over a small, predefined verb set, which struggles to generalize to the long-tail of unseen or ambiguous interactions in the wild. While recent multi-modal large language models (MLLMs) possess the rich world knowledge required for open-vocabulary understanding, they remain decoupled from existing HOI detectors since fine-tuning them is computationally prohibitive. To address these constraints, we propose \textbf{GRASP-HOI}, a novel \underline{G}enerative \underline{R}easoning \underline{A}nd \underline{S}teerable \underline{P}erception framework that reformulates HOI detection from the closed-set classification task to the open-vocabulary generation problem. To bridge the vision and cognitive, we first extract hybrid interaction representations, then design a lightweight learnable cognitive steering conduit (CSC) module to inject the fine-grained visual evidence into a frozen MLLM for effective reasoning. To address the supervision mismatch between classification-based HOI datasets and open-vocabulary generative models, we introduce a hybrid guidance strategy that coupling the language modeling loss and auxiliary classification loss, enabling discriminative grounding without sacrificing generative flexibility. Experiments demonstrate state-of-the-art closed-set performance and strong zero-shot generalization, achieving a unified paradigm that seamlessly bridges discriminative perception and generative reasoning for open-world HOI detection.
\vspace{-1mm}
\end{abstract}    
\vspace{-2mm}
\section{Introduction}
\label{sec:intro}

Human-object interaction (HOI) detection aims to localize human-object pairs and identify interactions connecting them, which is essential for applications including visual understanding \cite{kim2024TETAD, xiong2024Modalitycollaborative}, scene graph generation \cite{li2024Pixels, zhang2024HiKERSGG} and embodied artificial intelligence \cite{ma2025surveyvisionlanguageactionmodelsembodied, hoi4bot}\textit{etc.} While significant progresses have been made in traditional HOI detection \cite{han2025Survey}, these advancements are largely anchored to a closed-world classification paradigm. This dominant approach, which trains models with classification heads to select verbs from a small predefined vocabulary, has achieved remarkable performance on existing widely used datasets \cite{gkioxari2018detecting, cao2023Remine, yuan2022RLIP}. However, the reliance on a fixed label set constitutes a fundamental paradigmatic bottleneck, limiting their generalization capabilities on unseen, compositional, or ambiguous interactions in real-world scenarios.

\begin{figure}[t]
    \centering
    \includegraphics[width=1\linewidth]{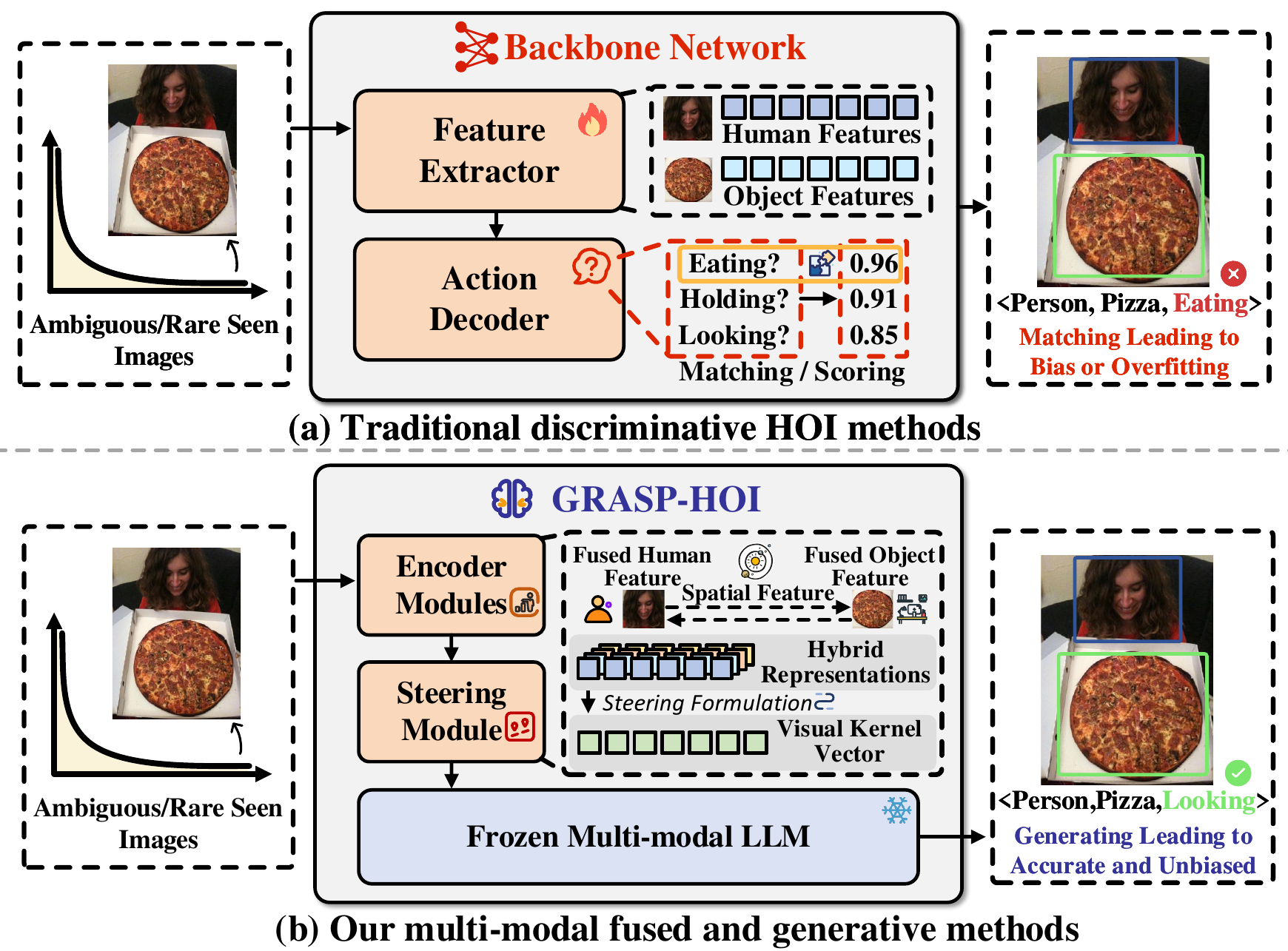}
    \caption{An overview of the traditional discriminative matching method and the proposed generative reasoning paradigm for HOI detection. (a) Traditional methods classify each detected human-object pair via classification or matching, limiting to frequent labels. (b) GRASP-HOI fuses multi-source features to steer a frozen MLLM, generating context-aware interactions beyond closed-set.}
    \label{fig:first}
    \vspace{-2mm}
\end{figure}

Recent advances in large language models (LLMs) \cite{touvron2023llama, yang2025qwen3technicalreport, achiam2023gpt} and multimodal large language models (MLLMs) \cite{bai2025Qwen25VL, liu2023Visual, zhu2025internvl3} have demonstrated impressive open-world reasoning capabilities. These models possess extensive world knowledge and linguistic abstraction that enable fine-grained understanding beyond fixed label spaces. However, leveraging such models for HOI remains unaddressed. Frozen MLLMs are generative in nature yet inaccurate and uncontrollable, while fine-tuning MLLMs on specific HOI datasets can lead better performance but requires substantial computational resources and even more risks catastrophic forgetting, degrading the world knowledge we seek to leverage \cite{zhang2025survey, luo2025empirical}. On the other hand, existing HOI detectors are highly discriminative but semantically rigid, unable to adapt beyond closed-set verb semantics \cite{cao2023Detecting}. This exposes a fundamental paradigm gap, \textit{i.e.}, current systems either detect without reasoning or reason without grounding.

Several attempts have sought to extend HOI detection toward open-vocabulary recognition by coupling visual encoders with pretrained text models \cite{ning2023HOICLIP, yang2024Openworld, wu2024Openset}. While these approaches enlarge the verb space through textual similarity or compositional prompts, they remain passive matchers rather than active generators, simply transferring the label bottleneck from visual categories to textual embeddings. Consequently, they still fail to express new interactions that require contextual or causal reasoning, such as catching a falling book or holding a phone to illuminate. We argue that this failure stems from their reliance on shallow correlation alignment rather a genuine comprehension with  the deep causal and functional reasoning. 

More recently, some studies have explored using LLMs or MLLMs as captioning assistants for detected entities \cite{hu2025Bilateral, luo2024Discovering, cao2023Detecting}. Although these efforts reflect a growing interest in generative understanding, they remain passive utilization of large models, treating them as external captioners or auxiliary classifiers detached from the visual understanding and detecting process \cite{cao2023Detecting, yang2024Openworld, lei2024Exploring}. Such designs fail to unlock the interactive potential of MLLMs, as they lack a mechanism to proactively utilize and steer them for human-object interaction detection. This leaves a critical blank spot, \textit{i.e.}, the absence of a framework that can actively utilize a frozen MLLM visual reasoning and generating results.

To address above constraints, we propose \textbf{GRASP-HOI}, (Generative Reasoning And Steerable Perception for HOI detection), a framework that transforms HOI detection from passive matching to  the active generative reasoning paradigm. Instead of tuning or prompting a large multimodal model, GRASP-HOI establishes a differentiable cognitive and knowledge interface between vision and language, which allows structured visual features to dynamically steer a frozen MLLM toward task-aligned reasoning. Specifically, GRASP-HOI first extracts hybrid interaction representations, then learns to convert interaction-centric visual representations into an internal cognitive state that modulates the MLLM’s generation trajectory while preserving its pretrained world knowledge. Through this synergy, with the help of language loss and classification loss, perception constrains cognition with pixel-level grounding, whereas cognition enriches perception with causal and functional semantics. The resulting system unifies the discriminative precision of classical detectors with the semantic generality of large generative models, enabling both robust closed-set performance and flexible open-vocabulary reasoning. By turning passive exploitation of MLLMs into active, trainable cognitive collaboration, GRASP-HOI advances a paradigm shift from detection-by-classification to reasoning-by-generation, bridging visual grounding and high-level conceptual understanding within a single, coherent framework. Our contributions are as follows:

\begin{itemize}

    \item We introduce GRASP-HOI, a novel framework that reframes HOI detection as precptual generation by steering a frozen MLLM with visual evidence, shifting from the closed-set classification to open world generation.

    \item We design the cognitive steering conduit (CSC), a novel hybrid guidance mechanism that establishes controllable generative reasoning between structured visual representations and a frozen MLLM.
    \item We develop a hybrid generative objective that couples language modeling, cross-modal alignment, and commonsense constraints with detection losses, ensuring generation grounded and evaluable.
    \item We conduct comprehensive experiments on HICO-DET and V-COCO, demonstrating state-of-the-art performance in standard benchmarks and substantial gains under zero-shot and open-vocabulary settings.
\end{itemize}

\section{Related Work}
\label{sec:relate}

\subsection{Traditional HOI detection}
Traditional HOI detectors typically follow either a two-stage or one-stage paradigm \cite{han2025Survey}. Two-stage methods first rely on object detector to localize humans and objects, and then construct interaction representations through multi-stream feature fusion, including appearance, spatial configuration, human pose, or other cues \cite{zhang2022Efficient, liu2020ConsNet, wang2024Bilateral}. To enhance contextual reasoning, several works further integrate graph neural networks to propagate relational information among detected entities \cite{park2023ViPLO, ren2024Learning}. In contrast, one-stage approaches unify detection and interaction recognition within a transformer-based decoder, such as DETR-style architectures, achieving greater efficiency and global context awareness \cite{yuan2023RLIPv2, ma2024FGAHOI, kim2021HOTR}. Despite these advances, both paradigms treat interactions as classification over a fixed verb set, thus inherently constrained by closed-world semantics. This limitation has motivated growing interest in open-vocabulary HOI detection.

\subsection{Open-vocabulary HOI Detection}
To overcome the restricted label space, recent studies explore open-vocabulary HOI recognition through three primary directions. The first category adopts compositional learning to generalize unseen $\langle\text{human}, \text{verb}, \text{object}\rangle$ combinations \cite{xue2025Zeroshot, yang2024Openworld}; the second emphasizes large-scale pretraining by exploiting external datasets or pseudo-labeled data to broaden the verb-object coverage \cite{zheng2023OpenCategory, yuan2022RLIP, li2024Disentangled}; and the third incorporates vision-language models (VLMs) for cross-modal alignment \cite{wu2023Endtoend, ning2023HOICLIP, wan2024exploiting}. Among the latter, some distill knowledge from VLMs into HOI detectors during training, while others directly fuse VLM features or textual embeddings to improve interaction recognition \cite{luo2024Discovering, zhan2024enhancinghoidetectioncontextual}. Although these methods expand the vocabulary and improve zero-shot matching, they still operate as passive matchers, relying on static text embeddings or similarity ranking rather than genuine generative reasoning. Consequently, they cannot synthesize novel or context-dependent interactions beyond predefined textual priors, calling for more active integration of foundational generative models.

\subsection{Foundational Model based HOI Detection}

The emergence of large language models (LLMs) \cite{vaswani2017Attention, qwen2025qwen25technicalreport} and multimodal large language models (MLLMs) \cite{zhang2023VideoLLaMA, zhu2025internvl3, bai2023QwenVL} has brought unprecedented capabilities in open-world reasoning and linguistic abstraction. Several recent works leverage such models via captioning pipelines or feature adapters to enhance interaction understanding \cite{hu2025Bilateral}. Representative frameworks demonstrate that foundational models can be guided for visual reasoning with lightweight adapters or prompts \cite{lei2025HOLa, kim2025Localityaware}. However, when applied to HOI detection, these methods remain non-differentiable or weakly coupled, lacking fine-grained visual grounding and controllable supervision. Most current systems still treat MLLMs as auxiliary captioners or feature extractors rather than actively co-reasoning with structured detectors. In contrast, our approach introduces a trainable, differentiable interface that enables active cognitive alignment between structured HOI perception and a frozen MLLM’s generative reasoning process, bridging the long-standing gap between discriminative detection and open-world understanding.
\section{Method}
\label{sec:method}

\begin{figure*}[t]
    \centering
    \includegraphics[width=1\linewidth]{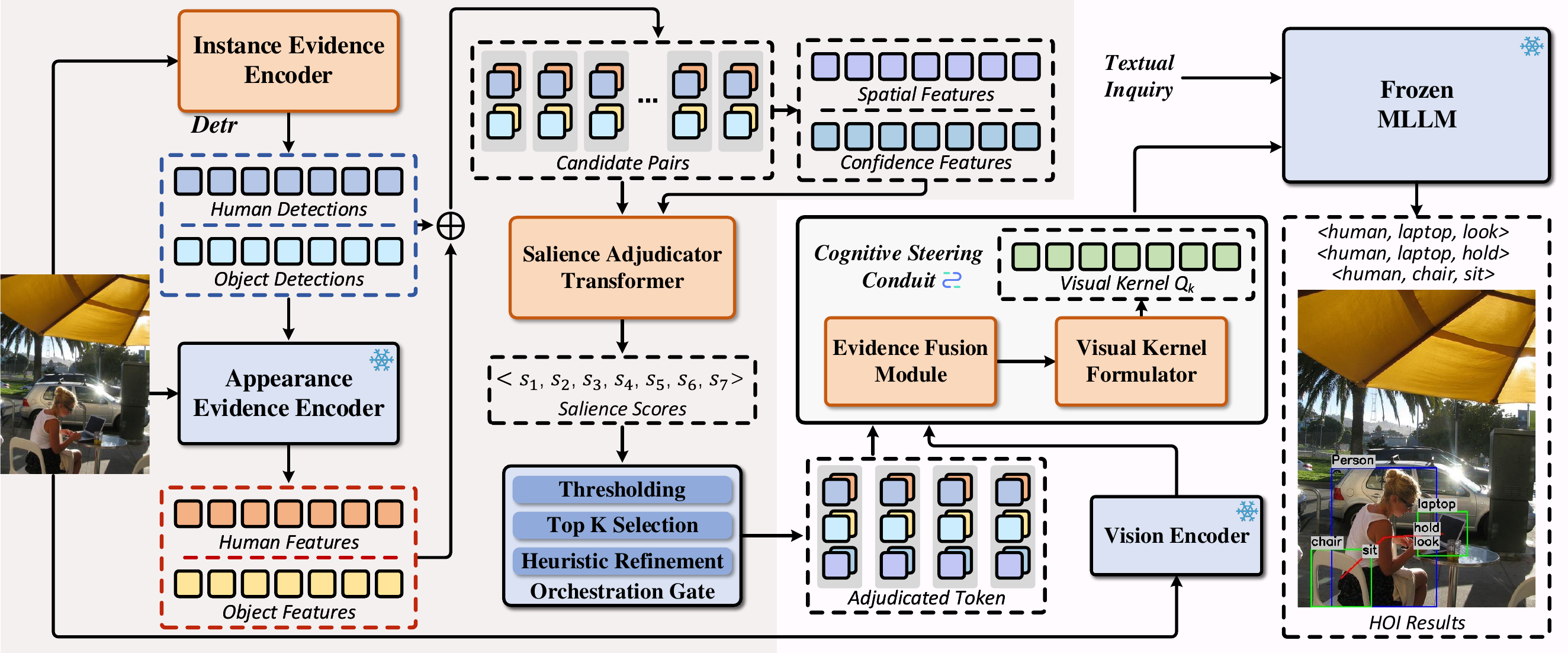}
    \caption{The architecture of GRASP-HOI, which performs open-vocabulary HOI detection by first process multi-source representations then steering a frozen generative model to describe them. The Instance Evidence Encoder and Appearance Evidence Encoder provide identified humans and objects in the image and extract visual features from the detected human, object, and their bounding box. Then a salience adjudication transformer and an Orchestration Gate distill the set of interaction features. The Cognitive Steering Conduit adjudicated candidate token with a global scene token from the frozen MLLM vision encoder into an evidence vector $e_k$. The visual kernel formulator transduces $e_k$ into a sequential visual kernel $Q_k$ to finally guide the frozen MLLM. This process enables GRASP-HOI to leverage a powerful, frozen MLLM for HOI detection with minimal, targeted training.}
    \label{fig:arch}
\end{figure*}

GRASP-HOI introduces a novel framework that reframes human-object interaction (HOI) detection as learnable generative process via the frozen multimodal large language model (MLLM). As shown in Figure~\ref{fig:arch}, we first construct the hybrid interaction representation that aggregates entity-level and pair-level visual evidence and use salience adjudication transformer with orchestration gate to organizes multi-source visual evidence into candidate-wise tokens. Then we perform cognitive steering process to formulate the representation into a compact evidence vector and transduces it into a visual kernel that conditions the frozen MLLM to generate potential interactions.

\subsection{Problem Formulation}
\label{sec:problem}

Given an image $I$, Human-Object Interaction (HOI) detection aims to predict a set of visually grounded triplets
\(
    \langle b_h, v, b_o \rangle
\),
where $b_h, b_o \in \mathbb{R}^4$ denote the bounding boxes of human and object, and $v \in \mathcal{V}$ is a verb from the verb set $\mathcal{V}$. We build on top of an off-the-shelf detector that provides human and object detections; these detections serve as anchors to form candidate human-object pairs. GRASP-HOI performs candidate-conditioned interaction reasoning by constructing candidate-level visual evidence and using a frozen multimodal large language model (MLLM) to generate verb predictions in a candidate-specific manner.

\subsection{Hybrid Interaction Representation}

The initial step of our framework is to construct a hybrid interaction representation with multi-source low-level visual signals. These representations are processed and provide rich substrate for cognitive steering.

\vspace{-4mm}
\paragraph{Entity-level Representations.}
For each detected entity $x$ (human or object), we form a canonical entity evidence by combining two complementary tokens including the instance token $z_x$ from a query-based detector and the appearance token $a_x$ obtained via RoI pooling from a vision backbone. The instance tokens encode detection-aligned semantics and localization (\emph{what \& where}) while the appearance tokens contribute fine-grained visual cues that are robust under distribution shift. We first project tokens to a unified dimension and then fuse them with a shallow MLP:
\begin{equation}
\label{eq:entity}
    f_x \;=\; \mathrm{MLP}_{\text{fuse}}\!\Big(\,\big[\,\phi_{\text{inst}}(z_x)\;\|\;\phi_{\text{app}}(a_x)\,\big]\,\Big)\;\in\;\mathbb{R}^{d_e},
\end{equation}
where $\phi_{\text{inst}}(\cdot)$ and $\phi_{\text{app}}(\cdot)$ are small projection MLPs and $[\,\cdot\|\cdot\,]$ denotes concatenation. The resulting $f_x$ is a detection-grounded, appearance-enriched evidence carrier that will be combined with pairwise geometry to construct candidate tokens for subsequent adjudication.

\vspace{-4mm}
\paragraph{Pair-Level Geometric Representations.} 
For a candidate pair $k$ with boxes $(b_h^k,b_o^k)$, we encode geometry as
\begin{equation}
\label{eq:geom}
    g_k = \phi_g\big(\mathcal{G}(b_h^k,b_o^k)\big) \in \mathbb{R}^{d_g},
\end{equation}
The function $\mathcal{G}$ computes a low-dimensional vector representing the geometric relationship between entities, including their normalized distance, scale, and overlap. The precise definition is deferred to the appendix.

\vspace{-4mm}
\paragraph{Synergistic Adjudication.}
Let $\mathcal{P} = \{(h_k,o_k)\}_{k=1}^N$ be the set of all human-object pairs. For each candidate $k$, we aggregate its evidence into a single token $u_k$ by concatenating their entity representations $(f_h,f_o)$ and geometry $g_k$, followed by linear projection to the SAT dimension $d_{\text{model}}$. The resulting sequence $\{u_k\}_{k=1}^N$ is fed into the salience adjudication transformer (SAT), a stack of transformer encoder layers with multi-head self-attention and feed-forward blocks. SAT performs scene-level contextual reasoning that each candidate attends to all others to capture patterns such as competition for the same object or mutually exclusive poses, and produces contextualized tokens $\tilde{u}_k$.

The shared linear head followed by a sigmoid maps each $\tilde{u}_k$ to a raw salience score
\begin{equation}
    s_k = \sigma(w_s^\top \tilde{u}_k + b_s),
\end{equation}
reflecting the model belief that candidate $k$ corresponds to a valid interaction. To incorporate low-level evidence quality, the Orchestration Gate (OG) combines $s_k$ with detector confidences $\text{conf}_h^k$ and $\text{conf}_o^k$ into a refined score
\begin{equation}
    r_k = \alpha\, s_k + (1-\alpha)\,\min(\text{conf}_h^k,\text{conf}_o^k),
\end{equation}
where $\alpha$ is a balancing hyperparameter. We then apply simple per-human quotas and coverage heuristics to select a compact candidate set $\mathcal{P}^\star$. For each $(h_k,o_k)\in\mathcal{P}^\star$, we denote its contextualized token as $v_k$ and refer to it as the adjudicated candidate token, which serves as structured visual evidence for the subsequent cognitive steering stage.

\subsection{Cognitive Steering Conduit}

The cognitive steering stage translates the adjudicated perceptual evidence into fine-grained instructions for the frozen MLLM. While The SAT token $\tilde{u}_k$ captures local interaction evidence and candidate-wise context, the vision tower from MLLM provides a global scene token capturing high-level semantics aligned with the text space. The Evidence Fusion Module aligns and compresses them into a compact evidence vector, then the visual kernel formulator unfolds this vector into a sequence of visual kernel tokens for precise control over the generative process.

\vspace{-4mm}
\paragraph{Evidence Fusion.}
For each selected candidate $(h_k,o_k)\in\mathcal{P}^\star$, we take its adjudicated token $v_k$ as the local interaction evidence. To complement this with global semantics aligned to the MLLM, we extract a global scene token $f_{\text{global}}\in\mathbb{R}^{D_g}$ from the frozen MLLM vision encoder by average pooling its last-layer patch features. Since $v_k$ and $f_{\text{global}}$ lie in heterogeneous feature spaces, we first project them to the MLLM hidden dimension $d$ and then fuse them with a shallow MLP:
\begin{equation}
\label{eq:evidence}
    e_k = \mathrm{MLP}_{\text{fuse}}\Big(\big[\;\phi_c(v_k)\;\|\;\phi_g(f_{\text{global}})\;\big]\Big)\in\mathbb{R}^{d},
\end{equation}
where $\phi_c(\cdot)$ and $\phi_g(\cdot)$ are small projection MLPs and $[\,\cdot\|\cdot\,]$ denotes concatenation. The resulting $e_k$ is a compact, candidate-specific evidence vector expressed directly in the MLLM latent space, which serves as the sole input to the subsequent visual kernel formulator.

\vspace{-4mm}
\paragraph{Visual Kernel Formulation.}
The visual kernel formulator (VKF) a lightweight cross-attention module, transduces the dense evidence vector $e_k$ into a structured, sequential condition for the MLLM. 

\begin{equation}
\label{eq:qk}
    Q_k = \mathrm{FFN}\big(\mathrm{MHCA}(Z,\,e_k)\big)\in\mathbb{R}^{L\times d}.
\end{equation}

$\mathrm{MHCA}$ uses $e_k$ as a shared memory (key/value) for all slots; $L$ controls prompt capacity. The resulting $Q_k$ is the candidate’s visual kernel that conditions the MLLM to reason about the specific interaction. 


\begin{figure}[t]
    \centering
    \includegraphics[width=1\linewidth]{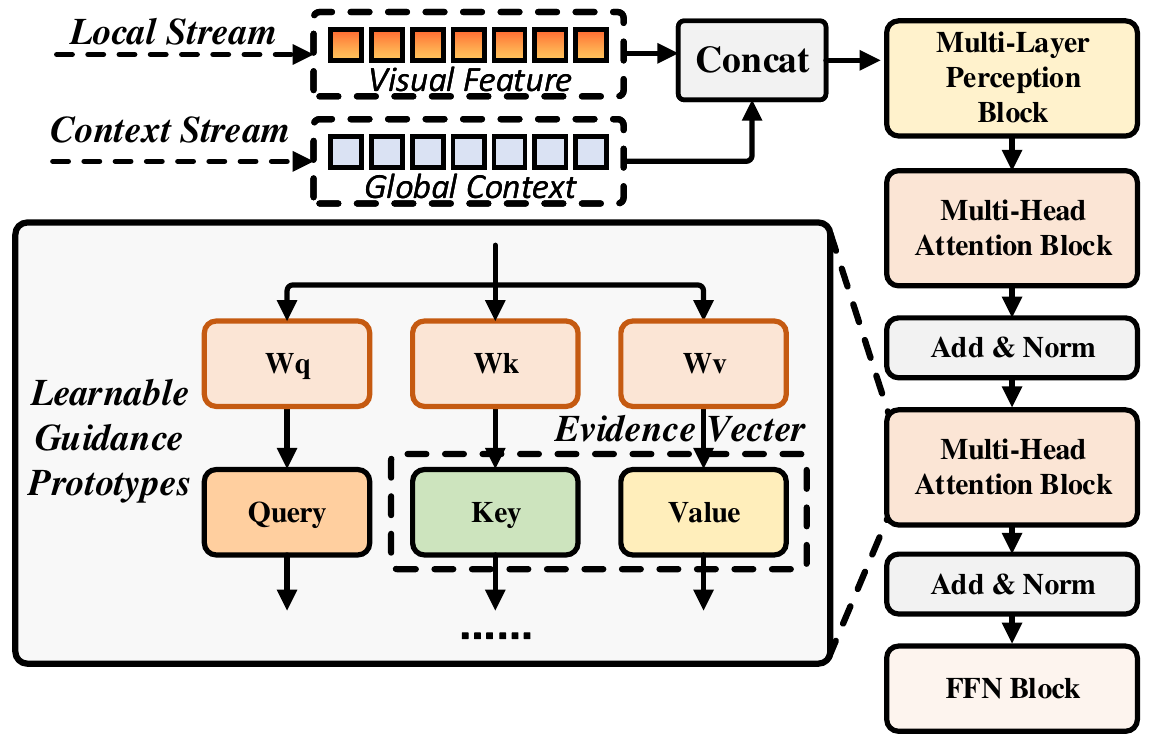}
    \caption{The architecture of Cognitive Steering Conduit. The evidence fusion module produces unified evidence vector $e_k$. The visual kernel formulator then transduces $e_k$ into the visual kernel $Q_k \in \mathbb{R}^{L \times d}$ which serves as a soft visual prefix to steer the frozen MLLM.}
    \label{fig:csc_detail}
\end{figure}

\vspace{-4mm}
\paragraph{Kernel-Conditioned Generation.}
The generated visual kernel $Q_k$ is prepended to the token embeddings of a textual inquiry to form a unified input sequence for the frozen MLLM. Let $E(\cdot)$ be the text embedding layer and $T$ tokens of inquiry text, the final input sequence is given by:

\begin{equation}
\label{eq:inputseq}
    \tilde{X}_k = [\,Q_k,\,E(\text{text})\,]\in\mathbb{R}^{(L+T)\times d}.
\end{equation}

Within the autoregressive process of MLLM, the attention mechanism ensures that every subsequently generated text token attends to the prefixed visual kernel. $Q_k$ thus acts as persistent visual condition, steering the generation process to elaborate on the specific visual hypothesis it represents, rather than answering generic question.

\subsection{Training Objective}
\label{sec:training}

We jointly optimize all learnable components under a unified multi-task objective, while keeping the MLLM and DINO encoders frozen. The overall loss is defined as
\begin{equation}
\label{eq:total_loss}
\begin{split}
    \mathcal{L} = {}& 
    \lambda_{\text{det}}\mathcal{L}_{\text{det}} +
    \lambda_{\text{sal}}\mathcal{L}_{\text{sal}} \\
    &+ \lambda_{\text{gen}}\mathcal{L}_{\text{gen}} +
    \lambda_{\text{nce}}\mathcal{L}_{\text{nce}} +
    \lambda_{\text{logic}}\mathcal{L}_{\text{logic}},
\end{split}
\end{equation}

where $\lambda_{\cdot}$ are weighting coefficients. $\mathcal{L}_{\text{det}}$ follows the standard set-based detection loss used in query-based detectors, and $\mathcal{L}_{\text{sal}}$ is a binary cross-entropy loss with Hungarian matching that trains the SAT to predict salient candidates.

\vspace{-4mm}
\paragraph{Generative Consistency $\mathcal{L}_{\text{gen}}$.} 
For a positive candidate, given its ground-truth verb phrase $y_{1:T}$, we enforce
\begin{equation}
    \mathcal{L}_{\text{gen}}=-\sum_{t=1}^{T}\log p(y_t\mid y_{<t},Q_k),
\end{equation}
with decoding restricted to a verb-focused vocabulary during training, ensuring that $Q_k$ genuinely controls the generated verb sequence.

\vspace{-4mm}
\paragraph{Semantic Alignment $\mathcal{L}_{\text{nce}}$.}
We align kernels and verbs inside the same MLLM text–visual embedding space. Let $\bar{Q}_k=\frac{1}{L}\sum_{\ell=1}^L Q_k^{(\ell)}$, $w^+$ be the frozen text embedding of the ground-truth verb and $\mathcal{W}^-$ negatives. We use InfoNCE:
\begin{equation}
\small
    \mathcal{L}_{\text{nce}}=-\log\frac{\exp(\mathrm{cos}(\bar{Q}_k,w^+)/\tau)}
    {\sum_{w\in\{w^+\}\cup\mathcal{W}^-}\exp(\mathrm{cos}(\bar{Q}_k,w)/\tau)}.
\end{equation}
This is analogous to CLIP-style alignment, but with the visual kernel serving as the visual side.

\vspace{-4mm}
\paragraph{Logical Consistency $\mathcal{L}_{\text{logic}}$.}
We inject commonsense via a soft mutual-exclusion regularizer. For mutually exclusive verb pairs $\mathcal{M}$ (e.g., \emph{sit on} vs. \emph{stand on}), let $p(v\mid Q_k)$ be the probability of generating the canonical prompt of $v$ under teacher forcing (using the first verb token’s softmax). Then
\begin{equation}
    \mathcal{L}_{\text{logic}}=\sum_{(v,v')\in\mathcal{M}}\min\big(p(v\mid Q_k),\,p(v'\mid Q_k)\big).
\end{equation}
This penalizes assigning high probabilities to logically incompatible verbs without modifying MLLM parameters.

\subsection{Real-time Inference with GRASP-HOI}
\label{sec:output}

\paragraph{Constrained Decoding.}
For each $(h_k,o_k)\in\mathcal{P}^\star$, the MLLM generates a short description under $Q_k$. To ensure reproducible evaluation, we use constrained decoding over the verb set $\mathcal{V}$ (and minimal auxiliaries) and apply a simple template-based rule to extract the main verb, yielding $\langle b_{h_k},v_k,b_{o_k}\rangle$. We find constrained decoding crucial to mitigate semantic drift and stabilize metrics.

\vspace{-4mm}
\paragraph{Open-Vocabulary Mapping.}
For open-vocabulary evaluation, we relax the constraint, mapping the generated phrase to canonical verb via cosine similarity in the frozen MLLM text embedding space (the same space used by $\mathcal{L}_{\text{nce}}$), optionally filtered by WordNet synonyms.

\section{Experiments}
\label{sec:exp}

\begin{table*}[t]
\centering
\small
\setlength{\tabcolsep}{4.2pt} 
\renewcommand{\arraystretch}{1.05}
\caption{Comparison with state-of-the-art methods on HICO-DET and V-COCO in the closed-set setting. We report mAP (\%) and sort methods by HICO-DET Default Full.}
\begin{tabular}{lccccccccc}
\toprule
\multirow{2}{*}{\textbf{Method}} & \multirow{2}{*}{\textbf{Backbone}} & 
\multicolumn{3}{c}{\textbf{HICO-DET Default}} & 
\multicolumn{3}{c}{\textbf{HICO-DET Known-Object}} & 
\multicolumn{2}{c}{\textbf{V-COCO}} \\
\cmidrule(lr){3-5} \cmidrule(lr){6-8} \cmidrule(lr){9-10}
& & Full & Rare & Non-Rare & Full & Rare & Non-Rare & $AP_{\text{role}}^{\#1}$ & $AP_{\text{role}}^{\#2}$ \\
\midrule
QPIC \cite{tamura2021QPIC} & ResNet-101 & 29.90 & 23.92 & 31.69 & 32.38 & 26.06 & 34.27 & 58.8 & 61.0 \\
CDN \cite{zhang2021mining} & ResNet-50 & 31.78 & 27.55 & 33.05 & 34.53 & 29.73 & 35.96 & 62.3 & 64.4 \\
UPT \cite{lei2023Efficient} & ResNet-101 & 32.62 & 28.62 & 33.81 & 36.08 & 31.41 & 37.47 & 61.3 & 67.1 \\
RLIP \cite{yuan2022RLIP} & ResNet-50 & 32.84 & 26.85 & 34.63 & -- & -- & -- & 61.9 & 64.2 \\
GEN-VLKT \cite{liao2022GENVLKT} & R50+CLIP & 33.75 & 29.25 & 35.10 & 36.78 & 32.75 & 37.99 & 62.4 & 64.7 \\
HOICLIP \cite{ning2023HOICLIP} & ResNet-50 & 34.69 & 31.12 & 35.74 & 37.61 & 34.47 & 38.54 & 63.5 & 64.8 \\
CLIP4HOI \cite{mao2023cliphoi} & ResNet-50 & 35.33 & 33.95 & 35.74 & 37.19 & 35.27 & 37.77 & -- & 66.3 \\
CQL \cite{xie2023Category} & ResNet-101 & 36.03 & 33.16 & 36.89 & 38.82 & 35.51 & 39.81 & 66.5 & 69.9 \\
DP-HOI \cite{li2024Disentangled} & ResNet-50 & 36.56 & 34.36 & 37.22 & 39.37 & 36.59 & 40.20 & 66.6 & -- \\
AGER \cite{tu2023agglomerative} & ResNet-50 & 36.75 & 33.53 & 37.71 & 39.84 & 35.58 & 40.23 & 65.7 & 69.7 \\
FGAHOI \cite{ma2024FGAHOI} & Swin-L & 37.18 & 30.71 & 39.11 & 38.93 & 31.93 & 41.02 & 60.5 & 61.2 \\
VIPLO \cite{park2023ViPLO} & R50+CLIP & 37.22 & 35.45 & 37.75 & 40.61 & 38.82 & 41.15 & 62.2 & 68.0 \\
RmLR \cite{cao2023Remine} & ResNet-101 & 37.41 & 28.81 & 39.97 & 38.69 & 31.27 & 40.91 & 64.2 & 70.2 \\
CMMP \cite{lei2024Exploringa} & ResNet-50 & 38.14 & 37.75 & 38.25 & -- & -- & -- & -- & 64.0 \\
ADA-CM \cite{lei2023Efficient} & R50+CLIP & 38.40 & 37.52 & 38.66 & -- & -- & -- & 58.6 & 64.0 \\
EZ-HOI \cite{lei2024EZHOI} & ResNet-50 & 38.61 & 37.70 & 38.89 & -- & -- & -- & 60.5 & 66.2 \\
BCOM \cite{wang2024Bilateral} & ResNet-50 & 39.34 & 39.90 & 39.17 & 42.24 & 42.86 & 42.05 & 65.8 & 69.9 \\
UniHOI-1 \cite{cao2023Detecting} & R101+VIT-L & 40.95 & 40.27 & 41.32 & 43.26 & 43.12 & 43.25 & 68.1 & 70.8 \\
DiffHOI \cite{yang2023Boosting} & Swin-L & 41.50 & 39.96 & 41.96 & 43.62 & 41.41 & 44.28 & 65.7 & 68.2 \\
BC-HOI \cite{hu2025Bilateral} & ResNet-50 & 43.01 & 45.76 & 42.18 & 45.35 & 47.94 & 44.57 & 68.2 & 70.1 \\
PVIC \cite{zhang2023exploring} & Swin-L & 44.32 & 44.61 & 44.24 & 47.81 & 48.38 & 47.64 & 64.1 & 70.2 \\
MP-HOI \cite{yang2024Openworld} & Swin-L+ViT & 44.53 & 44.48 & 44.55 & -- & -- & -- & 66.2 & 67.6 \\
SICHOI \cite{luo2024Discovering} & R101+VIT-L & 45.04 & 45.61 & 44.88 & 48.16 & 48.37 & 48.09 & 71.1 & 75.6 \\
RLIPv2 \cite{yuan2023RLIPv2} & Swin-L & 45.09 & 43.23 & 45.64 & -- & -- & -- & 72.1 & 74.1 \\
PAFR \cite{wu2024Exploring} & Swin-L & 46.01 & 46.74 & 45.80 & 49.50 & 50.59 & 49.18 & 63.0 & 68.7 \\
HORP \cite{geng2025HORP} & Swin-L+CLIP & 47.53 & 46.81 & 47.74 & 51.24 & 50.78 & 51.38 & 68.9 & 71.1 \\
\midrule
\rowcolor{gray!10} \textbf{GRASP-HOI (Ours)} & \textbf{ResNet-50+MLLM} & \textbf{48.02} & \textbf{48.15} & \textbf{48.09} & \textbf{51.57} & \textbf{51.53} & \textbf{51.61} & \textbf{72.5} & \textbf{76.2} \\
\bottomrule
\end{tabular}

\vspace{-2mm}
\label{tab:sota_comparison_closed_set}
\end{table*}

\subsection{Datasets and Evaluation Metrics}
\label{sec:datasets}

\paragraph{HICO-DET.}
HICO-DET \cite{hicodet} is a large-scale HOI detection benchmark containing 47776 images with 80 object categories, 117 verbs, and 600 interaction triplets. We follow the standard split, using 38118 images for training and 9658 for testing. Following prior works, we report mean average precision (mAP) under two evaluation settings: \emph{Default}, where undetected objects are treated as negatives, and \emph{Known Object}, where ground-truth object boxes are given. For both settings, we further break down mAP into Full, Rare, and Non-rare categories, where Rare triplets have fewer than 10 training instances.

\paragraph{V-COCO.}
V-COCO \cite{vcoco} is built upon MS-COCO and contains 10346 images annotated with 26 action classes. We adopt the official dataset splits and follow the standard evaluation protocol using $\text{AP}^{\text{role}}_1$ and $\text{AP}^{\text{role}}_2$. A prediction is counted as correct only if both the human and object boxes and the action label are correct; $\text{AP}^{\text{role}}_2$ further considers cases with multiple objects per action.

\vspace{-4mm}
\paragraph{Zero-shot and Open-vocabulary Evaluation.}
To evaluate open-vocabulary performance on HICO-DET, we follow the widely adopted zero-shot protocols including (i) \emph{Unseen Composition} (UC), where a subset of $\langle\text{human},\text{verb},\text{object}\rangle$ triplets are held out while their verbs and objects remain seen; (ii) \emph{Unseen Object} (UO), where some object categories never appear during training; and (iii) \emph{Unseen Verb} (UV), where certain verbs are unseen. For each setting, we report mAP on the unseen, seen, and full splits.

\vspace{-1mm}
\subsection{Implementation Details}
\label{sec:impl}

GRASP-HOI is built on query-based detector (DETR) for human/object proposals and a frozen vision backbone (DINOv3) for appearance tokens. We used InternVL3.5 as the frozen MLLM backbone of our framework. We set $L$, $K$, $\alpha$, and $(\lambda_{\text{gen}},\,\lambda_{\text{nce}},\,\lambda_{\text{logic}})$ to $8$, $3$, $0.6$, and $(1.0,\,0.5,\,0.1)$, respectively. We train GRASP-HOI using AdamW with a weight decay of $10^{-4}$ and a cosine learning rate schedule. All experiments are conducted with a batch size of 16 on 8 NVIDIA A6000 GPUs.

\begin{table*}[t]
\centering
\small
\setlength{\tabcolsep}{4.8pt}
\renewcommand{\arraystretch}{1.05}
\caption{Open-vocabulary evaluation on HICO-DET. Performance is measured in mAP under four zero-shot settings: Rare First Unseen Combination (RF-UC), Non-rare First Unseen Combination (NF-UC), Unseen Object (UO), and Unseen Verb (UV).}
\begin{tabular}{l|ccc|ccc|ccc|ccc}
\toprule
\multirow{2}{*}{\textbf{Method}} &
\multicolumn{3}{c|}{\textbf{RF-UC}} &
\multicolumn{3}{c|}{\textbf{NF-UC}} &
\multicolumn{3}{c|}{\textbf{UO}} &
\multicolumn{3}{c}{\textbf{UV}} \\
& Unseen & Seen & Full & Unseen & Seen & Full & Unseen & Seen & Full & Unseen & Seen & Full \\
\midrule
GEN-VLKT \cite{liao2022GENVLKT} & 21.36 & 32.91 & 30.56 & 25.05 & 23.38 & 23.71 & 10.51 & 28.92 & 25.63 & 20.96 & 30.23 & 28.74 \\
RLIPv2-ParSeDA\cite{liao2022GENVLKT} & 21.45 & 35.85 & 32.97 & 22.81 & 29.52 & 28.18 & -- & -- & -- & -- & -- & -- \\
HOICLIP \cite{ning2023HOICLIP} & 25.53 & 34.85 & 32.99 & 26.39 & 28.10 & 27.75 & 16.20 & 30.99 & 28.53 & 24.30 & 32.19 & 31.09 \\
DP-HOI \cite{li2024Disentangled} & 30.49 & 36.17 & 35.03 & 28.87 & 29.98 & 29.76 & -- & -- & -- & 26.30 & 34.49 & 33.34 \\
DiffHOI \cite{yang2023Boosting} & 32.06 & 36.77 & 35.89 & -- & -- & -- & -- & -- & -- & -- & -- & -- \\
BCOM \cite{wang2024Bilateral} & -- & -- & -- & 33.12 & 31.76 & 32.03 & -- & -- & -- & -- & -- & -- \\
HOIGen \cite{guo2024unseen} & -- & -- & -- & 33.98 & 32.86 & 33.08 & 36.35 & 32.90 & 33.48 & 20.27 & 34.31 & 32.34 \\
UniHOI (BLIP2) \cite{cao2023Detecting} & 28.68 & 33.16 & 32.27 & 28.45 & 32.63 & 31.79 & 19.72 & 34.76 & 31.56 & 26.05 & 36.78 & 34.68 \\
EZ-HOI \cite{lei2024EZHOI} & 34.24 & 37.35 & 36.73 & -- & -- & -- & 38.17 & 36.02 & 36.38 & 28.82 & 38.15 & 36.84 \\
CMMP \cite{lei2024Exploringa} & 35.98 & 37.42 & 37.13 & 33.52 & 35.53 & 35.13 & 39.67 & 36.15 & 36.74 & 30.84 & 37.28 & 36.38 \\
SICHOI \cite{luo2024Discovering} & 34.24 & 41.58 & 40.11 & 34.52 & 36.06 & 35.75 & -- & -- & -- & -- & -- & -- \\
BC-HOI \cite{hu2025Bilateral} & 42.31 & 40.67 & 40.99 & 33.01 & 37.24 & 36.40 & 19.94 & 37.03 & 34.18 & 31.18 & 41.31 & 39.89 \\
\midrule
\rowcolor{gray!10}\shortstack{\textbf{GRASP-HOI (Ours)}} & \textbf{43.67} & \textbf{42.98} & \textbf{42.46} & \textbf{35.61} & \textbf{38.95} & \textbf{38.24} & \textbf{39.98} & \textbf{38.15} & \textbf{37.69} & \textbf{32.45} & \textbf{42.57} & \textbf{40.14} \\
\bottomrule
\end{tabular}

\label{tab:open_vocabulary_comparison}
\end{table*}

\subsection{Comparison with State-of-the-Arts}

\paragraph{Closed-Set Performance.}
We first compare GRASP-HOI with state-of-the-art (SOTA) methods on the standard HICO-DET and V-COCO datasets, with results summarized in Table~\ref{tab:sota_comparison_closed_set}. On the challenging HICO-DET default setting, our method establishes a new SOTA, achieving 48.02 mAP in full split, surpassing the previous leading method, HORP (47.53 mAP) by +0.49 mAP.
Notably, the capability of our generative framework is revealed in the long-tail rare split. GRASP-HOI achieves 48.15 mAP, establishing +1.34 mAP lead over HORP. While many recent methods achieve gains primarily by optimizing for the non-rare set, the improvement on the data-sparse rare set provides compelling evidence for our approach. It validates that our cognitive steering paradigm by effectively leveraging the rich, implicit world knowledge of the frozen MLLM, successfully breaks away from the spurious correlations that plague traditional classifiers in the long-tail distribution. This robust performance is also shown on V-COCO dataset, where GRASP-HOI also achieves state-of-the-art performance in $AP_{\text{role}}^{\#1}$ and $AP_{\text{role}}^{\#2}$ respectively, confirming the broad applicability and robustness of our framework.

\vspace{-4mm}
\paragraph{Open-Vocabulary Performance.}
We evaluate the central hypothesis of our work with the model capacity to generalize to unseen interactions, which is a fundamental limitation of closed-world detectors. Table~\ref{tab:open_vocabulary_comparison} presents the comprehensive evaluation across the four standard zero-shot settings on HICO-DET.
The most challenging setting is the unseen verb (UV) split, as it directly tests the ability to reason about and generate novel action semantics not present in the training data. In this challenging split, GRASP-HOI achieves 32.45 mAP on unseen verbs, outperforming the strong baseline BC-HOI (31.18 mAP) by +1.27 mAP. This result quantitatively demonstrates that our generative reformulation effectively breaks the closed-world assumption, producing valid inferences for verbs unseen during training. 
Also the superior generalization extends across all settings. In rare first unseen combination, our model achieves 43.67 mAP, surpassing BC-HOI by +1.36 mAP. We observe similar SOTA results in the NF-UC (35.61 mAP) and UO (39.98 mAP) settings. The consistent dominance across all four settings underscores the fundamental advantage of our generative reasoning paradigm, moving beyond simple matching to achieve true open-vocabulary understanding.

\subsection{Ablation Studies}
\label{sec:ablation}

We perform series of ablation studies on HICO-DET with default full setting to dissect the GRASP-HOI framework and quantify each module contribution. Our analysis is structured to validate three central hypotheses: first, the superiority of our generative paradigm over traditional classification; second, the critical role of the cognitive steering conduit (CSC) as the core steering mechanism; and third, the synergistic effect of our hybrid guidance losses.

\vspace{-4mm}
\paragraph{Analysis of Core Components.}
We validate our design via incremental build-up study in Table~\ref{tab:ablation_main}. We start with a non-generative baseline classifier, using the SAT output with a standard classification head, which yields 31.8 mAP. 
Replacing the classifier with our generative framework and using only the semantic alignment loss ($\mathcal{L}_{\text{nce}}$) boosts performance to 43.7 mAP (+11.9), confirming the benefit of leveraging the MLLM latent space. 
Next, introducing the generative consistency loss ($\mathcal{L}_{\text{gen}}$) to force autoregressive verb prediction adds another +2.8 mAP, demonstrating that fine-grained generative supervision is critical for accuracy.
Incorporating our full cognitive steering conduit (CSC) provides the largest gain of +1.52 mAP, identifying the CSC as the principal driver of performance. This confirms our designation that visual evidence is structured for the MLLM is as critical as the evidence itself.
Finally, adding the $\mathcal{L}_{\text{logic}}$ loss provides final +0.32 mAP refinement, bringing the model to its peak performance of 48.02 mAP by regularizing against commonsense-violating predictions.

\begin{table}[t]
\centering
\small
\setlength{\tabcolsep}{5pt}
\renewcommand{\arraystretch}{1.1}
\caption{
Ablation studies of core components of GRASP-HOI on HICO-DET
(Default Full). $\mathcal{L}_{\text{nce}}$, $\mathcal{L}_{\text{gen}}$ and CSC
denote the alignment loss, generative loss, and cognitive steering conduit.
}

\begin{adjustbox}{width=\linewidth, center}
\begin{tabular}{cccc|c}
\toprule
\textbf{Method} & $\mathcal{L}_{\text{nce}}$ & $\mathcal{L}_{\text{gen}}$ & \textbf{CSC} & \textbf{mAP} \\
\midrule
Baseline (Classifier) & - & - & - & 31.8 \\
+ Alignment Loss & \checkmark & - & - & 43.7 \\
+ Generative Loss & \checkmark & \checkmark & - & 46.5 \\
+ CSC (Full Model w/o $\mathcal{L}_{\text{logic}}$) & \checkmark & \checkmark & \checkmark & 47.7 \\
\rowcolor{gray!10} \textbf{GRASP-HOI (Full)} & \checkmark & \checkmark & \checkmark & \textbf{48.02} \\
\bottomrule
\end{tabular}
\end{adjustbox}
\label{tab:ablation_main}
\end{table}

\begin{table}[t]
\centering
\small
\setlength{\tabcolsep}{8pt}
\renewcommand{\arraystretch}{1.1}
\caption{
Ablation studies of the key components in cognitive steering conduit on HICO-DET (Default Full).
}

\begin{tabular}{l|c}
\toprule
\textbf{CSC Configuration} & \textbf{mAP} \\
\midrule
\rowcolor{gray!10} \textbf{Full Model (Local + Global + VKF)} & \textbf{48.02} \\
\midrule
w/o Global Evidence ($f_{\text{global}}$) & 46.1 \\
w/o Local Evidence ($v_k$) & 41.3 \\
w/ Naive Fusion (replace VKF with MLP) & 44.6 \\
\bottomrule
\end{tabular}

\label{tab:ablation_csc}
\end{table}

\begin{table}[t]
\centering
\small
\setlength{\tabcolsep}{8pt}
\renewcommand{\arraystretch}{1.1}
\caption{
Effect of the visual kernel length $L$ in the CSC on HICO-DET
(Default Full). Vary $L$ from 1 to 16 and report mAP (\%).
}

\begin{tabular}{l|cccc}
\toprule
\textbf{Kernel Length ($L$)} & 1 & 4 & 8 & 16 \\
\midrule
\textbf{HICO-DET (mAP)} & 45.9 & 47.3 & \textbf{48.02} & 48.01 \\
\bottomrule
\end{tabular}

\vspace{-3mm}
\label{tab:ablation_length}
\end{table}

\vspace{-4mm}
\paragraph{Dissection of the Cognitive Steering Conduit.}
Having established the CSC as the principal performance driver, we now dissect its internal architecture in Table~\ref{tab:ablation_csc} to isolate the contribution of each component. Removing the global scene token $f_{\text{global}}$ yields a notable 1.9 mAP reduction, confirming that holistic scene cues remain indispensable for resolving ambiguous or multi-object interactions.
Eliminating the local adjudicated token $v_k$ leads to a dramatic collapse to 41.3 mAP, underscoring that specific multi-source features is the core steering signal that enables precise reasoning within the MLLM.
Finally, substituting our visual kernel formulator (VKF) with a naive MLP projection further degrades performance to 44.6 mAP. This degradation highlights a key insight that simple feature projection cannot faithfully expose the perceptual structure required for generation, whereas the cross-attention design is critical for grounding and shaping an effective generative condition.

\vspace{-4mm}
\paragraph{Hyperparameter Analysis of Visual Kernel Length.}
We further examine the sensitivity to the kernel length $L$, a key hyperparameter of the CSC, as shown in Table~\ref{tab:ablation_length}. Performance peaks at $L{=}8$. Shorter kernels ($L{\le}4$) provide insufficient capacity to encode fine-grained visual cues, resulting in noticeable degradation. In contrast, extending the kernel to $L{=}16$ yields marginal gains while introducing additional computation. These results indicate that $L{=}8$ offers the most favorable balance between representational expressiveness and parameter efficiency, serving as the optimal configuration for stable generation.

\begin{figure}[t]
    \centering
    \begin{subfigure}[b]{0.49\columnwidth}
        \centering
        \includegraphics[width=\textwidth]{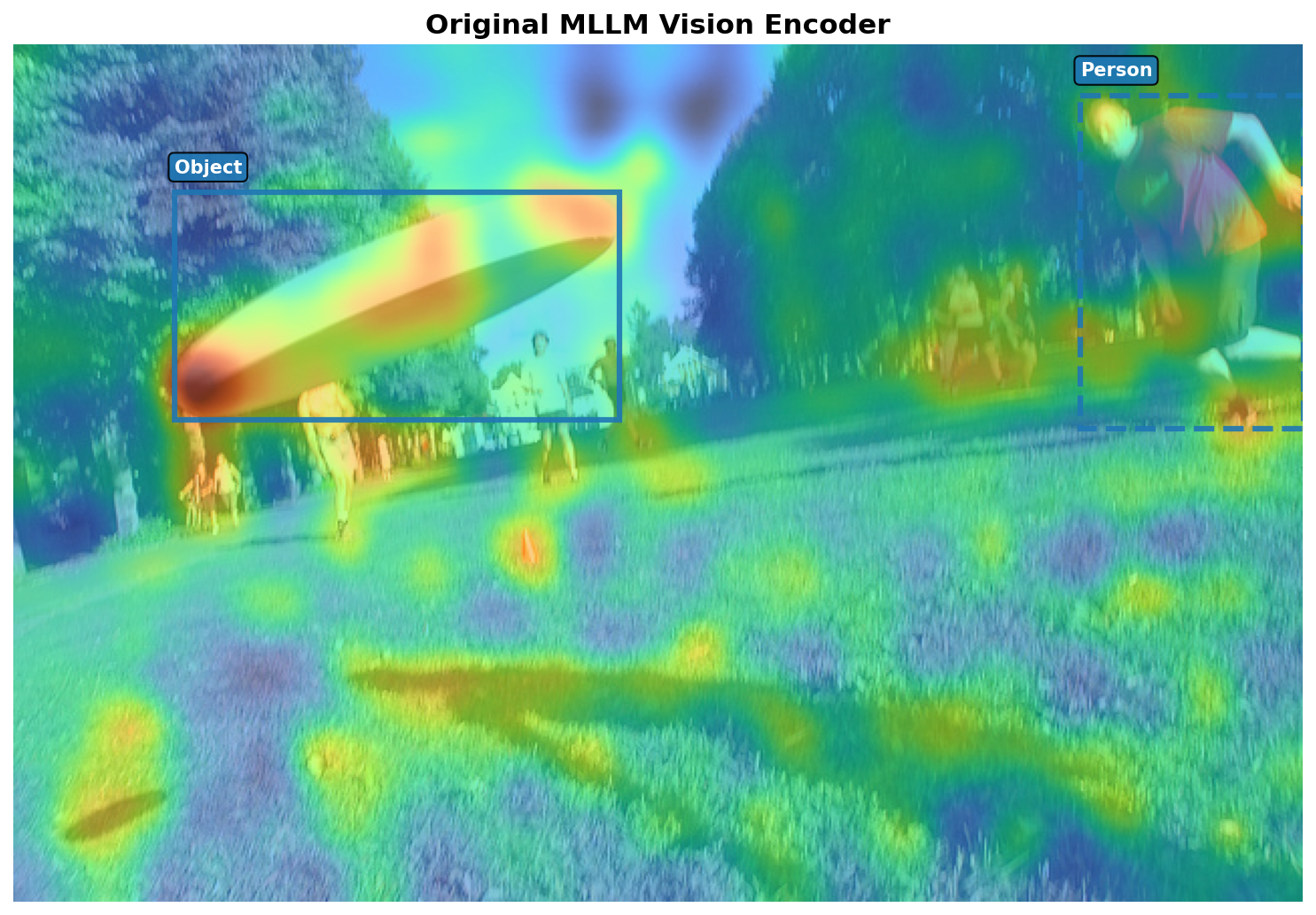}
    \end{subfigure}
    \begin{subfigure}[b]{0.49\columnwidth}
        \centering
        \includegraphics[width=\textwidth]{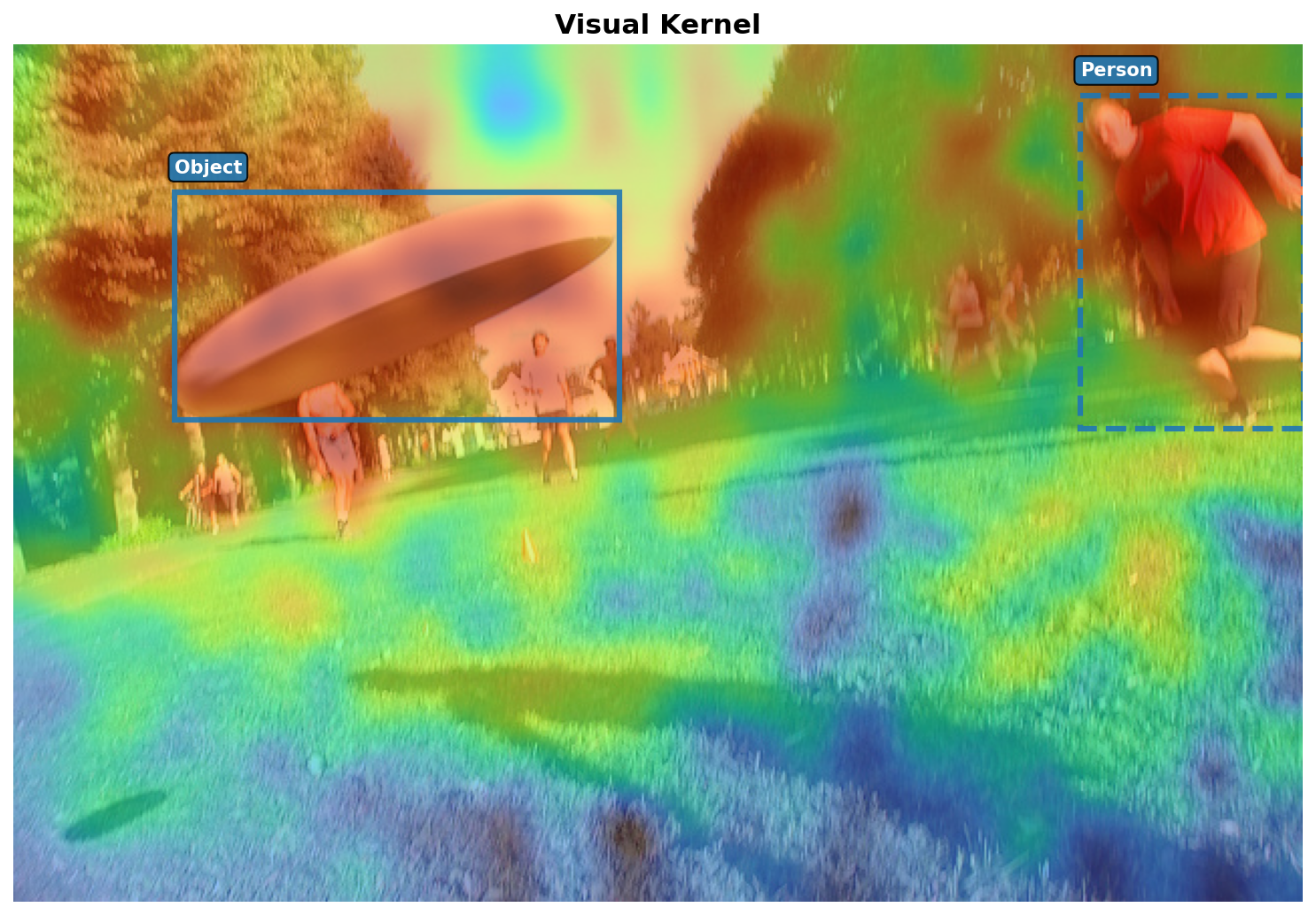}
    \end{subfigure}

    \vspace{-5mm}

    \begin{subfigure}[b]{0.49\columnwidth}
        \centering
        \includegraphics[width=\textwidth]{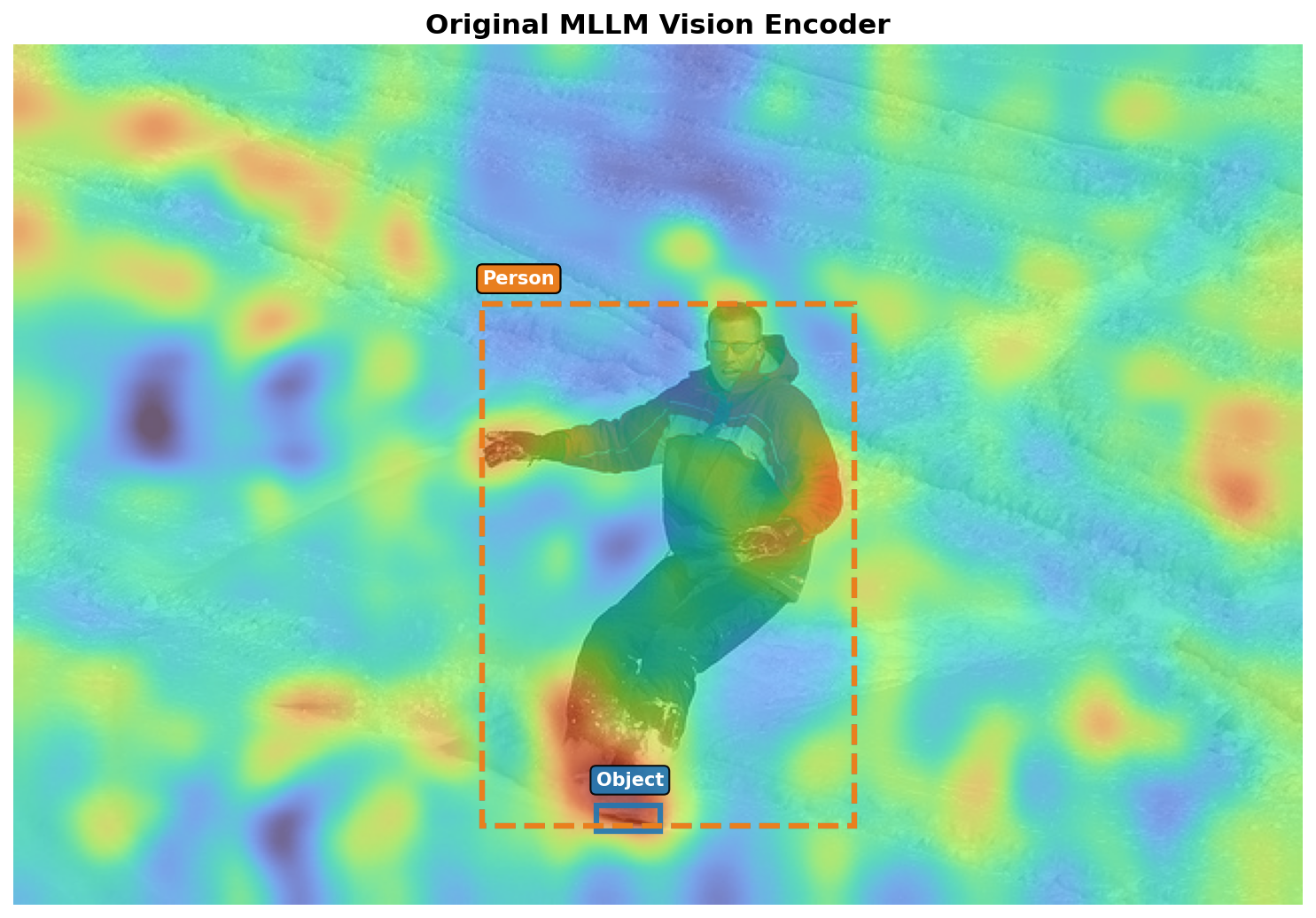}
        \caption*{(a)}
    \end{subfigure}
    \begin{subfigure}[b]{0.49\columnwidth}
        \centering
        \includegraphics[width=\textwidth]{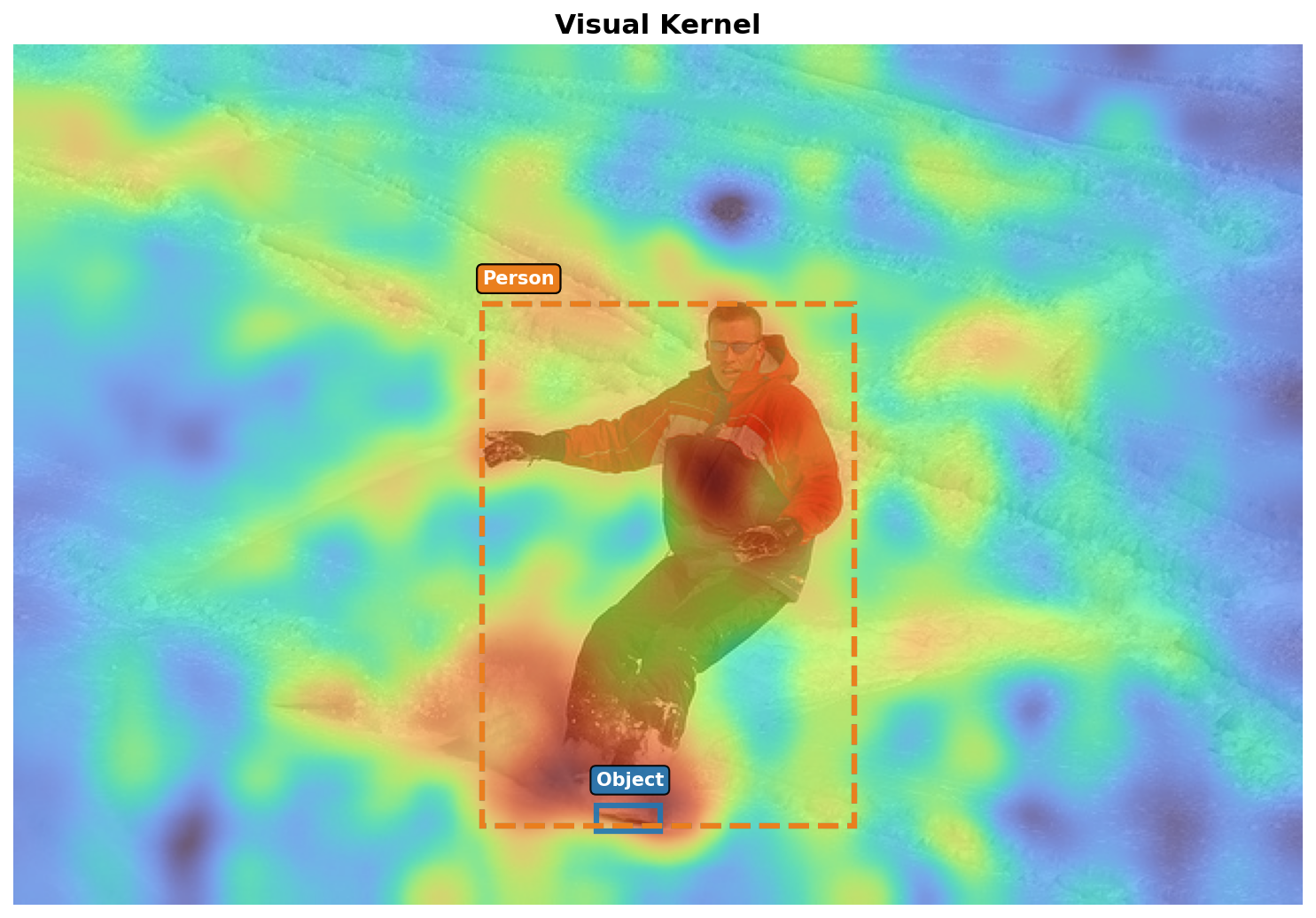}
        \caption*{(b)}
    \end{subfigure}

    \vspace{-3mm}
    \caption{
    Qualitative visualization of the steering effect of the Cognitive
    Steering Conduit (CSC) on HICO-DET. (a) Attention from the
    frozen vision encoder is diffuse and often focuses on irrelevant
    regions. (b) Our visual kernel yields concentrated responses on
    the target human-object interaction regions.
    }
    \vspace{-3.5mm}
    \label{fig:qualitative}
\end{figure}

\subsection{Qualitative Analysis}
\label{sec:qualitative}

We present qualitative visualizations of attention maps in Figure~\ref{fig:qualitative} to compare the responses of the frozen vision encoder in MLLM with produced after injecting candidate-specific visual kernel. The frozen encoder generally distributes attention broadly across background regions or non-interacting entities, reflecting its generic pre-training objectives. After applying visual kernel, the attention becomes sharply concentrated around the target human-object pair, highlighting both the interaction region and the supporting visual cues required for correct reasoning. This focused pattern demonstrates the visual kernel functions as an effective steering signal, grounding the MLLM’s high-level representations in the relevant perceptual evidence and reasoning capability. These visualizations provide intuitive confirmation that CSC enables the model to align generative reasoning with fine-grained, interaction-centric visual context.

\vspace{-1mm}
\section{Conclusion}
\label{sec:conclusion}
\vspace{-1 mm}
In this paper, we present GRASP-HOI, a framework that reformulates human-object interaction (HOI) detection from closed-set classification to open-vocabulary generative reasoning. At the core of our approach is the cognitive steering conduit (CSC), a lightweight and differentiable module that injects structured, candidate-specific visual evidence to steer a frozen multimodal large language model (MLLM). Enabled by a hybrid guidance strategy, the CSC leverages the MLLM’s reasoning capability without requiring any model fine-tuning. Extensive experiments show that GRASP-HOI achieves new state-of-the-art performance in both standard closed-set and open-vocabulary settings, demonstrating a superior and efficient paradigm that effectively bridges the long-standing gap between discriminative perception and high-level cognitive reasoning.

{
    \small
    \bibliographystyle{ieeenat_fullname}
    \bibliography{ref}

@String(CVPR= {IEEE Conf. Comput. Vis. Pattern Recog.})

@String(ICCV= {Int. Conf. Comput. Vis.})

@String(ECCV= {Eur. Conf. Comput. Vis.})

@String(AAAI = {AAAI})

@String(CVPR  = {CVPR})

@String(ICCV  = {ICCV})

@String(ECCV  = {ECCV})

@article{bai2023QwenVL,
  title={Qwen-VL: A Versatile Vision-Language Model for Understanding, Localization, Text Reading, and Beyond},
  author={Bai, Jinze and Bai, Shuai and Yang, Shusheng and Wang, Shijie and Tan, Sinan and Wang, Peng and Lin, Junyang and Zhou, Chang and Zhou, Jingren},
  journal={arXiv preprint arXiv:2308.12966},
  year={2023}
}

@inproceedings{cao2023Detecting,
  title = {Detecting any human-object interaction relationship: universal HOI detector with spatial prompt learning on foundation models},
  booktitle = {Proceedings of the Advances in Neural Information Processing Systems (NeurIPS)},
  author = {Cao, Yichao and Tang, Qingfei and Su, Xiu},
  year = 2023
}

@inproceedings{tu2023agglomerative,
  title={Agglomerative transformer for human-object interaction detection},
  author={Tu, Danyang and Sun, Wei and Zhai, Guangtao and Shen, Wei},
  booktitle={Proceedings of the IEEE/CVF International Conference on Computer Vision (ICCV)},
  pages={21614--21624},
  year={2023}
}

@inproceedings{mao2023cliphoi,
title={{CLIP}4{HOI}: Towards Adapting {CLIP} for Practical Zero-Shot {HOI} Detection},
author={Yunyao Mao and Jiajun Deng and Wengang Zhou and Li Li and Yao Fang and Houqiang Li},
booktitle={Proceedings of the Advances in Neural Information Processing Systems (NeurIPS)},
year={2023},
url={https://openreview.net/forum?id=nqIIWnwe73}
}

@inproceedings{geng2025HORP,
  title = {HORP: human-object relation priors guided HOI detection},
  booktitle = {Proceedings of the IEEE/CVF Conference on Computer Vision and Pattern Recognition (CVPR)},
  author = {Geng, Pei and Yang, Jian and Zhang, Shanshan},
  year = 2025
}

@article{vcoco,
  title={Visual semantic role labeling},
  author={Gupta, Saurabh and Malik, Jitendra},
  journal={arXiv preprint arXiv:1505.04474},
  year={2015}
}

@article{han2025Survey,
  title = {A Survey of Human-Object Interaction Detection With Deep Learning},
  author = {Han, Geng and Zhao, Jiachen and Zhang, Lele and Deng, Fang},
  year = 2025,
  journal = {IEEE Transactions on Emerging Topics in Computational Intelligence},
  volume = {9},
  number = {1},
  pages = {3--26},
  doi = {10.1109/TETCI.2024.3518613}
}

@inproceedings{gkioxari2018detecting,
  title={Detecting and recognizing human-object interactions},
  author={Gkioxari, Georgia and Girshick, Ross and Doll{\'a}r, Piotr and He, Kaiming},
  booktitle={Proceedings of the IEEE conference on computer vision and pattern recognition},
  pages={8359--8367},
  year={2018}
}

@inproceedings{hu2025Bilateral,
  title={Bilateral Collaboration with Large Vision-Language Models for Open Vocabulary Human-Object Interaction Detection},
  author={Hu, Yupeng and Ding, Changxing and Sun, Chang and Huang, Shaoli and Xu, Xiangmin},
  booktitle={Proceedings of the IEEE/CVF International Conference on Computer Vision},
  pages={20126--20136},
  year={2025}
}

@inproceedings{kim2021HOTR,
  title = {HOTR: End-to-End Human-Object Interaction Detection with Transformers},
  shorttitle = {Hotr},
  booktitle = {Proceedings of the IEEE/CVF Conference on Computer Vision and Pattern Recognition (CVPR)},
  author = {Kim, Bumsoo and Lee, Junhyun and Kang, Jaewoo and Kim, Eun-Sol and Kim, Hyunwoo J.},
  year = 2021,
  pages = {74--83},
  doi = {10.1109/CVPR46437.2021.00014}
}

@inproceedings{kim2024TETAD,
  title = {TE-TAD: towards full end-to-end temporal action detection via time-aligned coordinate expression},
  shorttitle = {Te-tad},
  booktitle = {Proceedings of the IEEE/CVF Conference on Computer Vision and Pattern Recognition (CVPR)},
  author = {Kim, Ho-Joong and Hong, Jung-Ho and Kong, Heejo and Lee, Seong-Whan},
  year = 2024,
  pages = {18837--18846},
  doi = {10.1109/CVPR52733.2024.01782}
}

@inproceedings{kim2025Localityaware,
  title={Locality-Aware Zero-Shot Human-Object Interaction Detection},
  author={Kim, Sanghyun and Jung, Deunsol and Cho, Minsu},
  booktitle={Proceedings of the IEEE/CVF Conference on Computer Vision and Pattern Recognition (CVPR)},
  pages={20190--20200},
  year={2025}
}

@inproceedings{lei2023Efficient,
  title = {Efficient adaptive human-object interaction detection with concept-guided memory},
  booktitle = {Proceedings of the IEEE/CVF International Conference on Computer Vision (ICCV)},
  author = {Lei, Ting and Caba, Fabian and Chen, Qingchao and Jin, Hailin and Peng, Yuxin and Liu, Yang},
  year = 2023,
  pages = {6457--6467},
  doi = {10.1109/ICCV51070.2023.00596}
}

@inproceedings{lei2024Exploring,
  title = {Exploring the potential of large foundation models for open-vocabulary HOI detection},
  booktitle = {Proceedings of the IEEE/CVF Conference on Computer Vision and Pattern Recognition (CVPR)},
  author = {Lei, Ting and Yin, Shaofeng and Liu, Yang},
  year = 2024,
  eprint = {2404.06194},
  primaryclass = {cs},
  doi = {10.48550/arXiv.2404.06194},
  archiveprefix = {arXiv}
}

@inproceedings{lei2024Exploringa,
  title={Exploring conditional multi-modal prompts for zero-shot hoi detection},
  author={Lei, Ting and Yin, Shaofeng and Peng, Yuxin and Liu, Yang},
  booktitle={Proceedings of the European Conference on Computer Vision (ECCV)},
  pages={1--19},
  year={2024},
  organization={Springer}
}

@article{lei2024EZHOI,
  title={Ez-hoi: Vlm adaptation via guided prompt learning for zero-shot hoi detection},
  author={Lei, Qinqian and Wang, Bo and Tan, Robby},
  journal={Proceedings of the Advances in Neural Information Processing Systems (NeurIPS)},
  volume={37},
  pages={55831--55857},
  year={2024}
}

@inproceedings{lei2025HOLa,
  title={HOLa: Zero-Shot HOI Detection with Low-Rank Decomposed VLM Feature Adaptation},
  author={Lei, Qinqian and Wang, Bo and Tan, Robby T},
  booktitle={Proceedings of the IEEE/CVF International Conference on Computer Vision (ICCV)},
  pages={1825--1835},
  year={2025}
}

@inproceedings{li2024Disentangled,
  title = {Disentangled Pre-Training for Human-Object Interaction Detection},
  booktitle = {Proceedings of the IEEE/CVF Conference on Computer Vision and Pattern Recognition (CVPR)},
  author = {Li, Zhuolong and Li, Xingao and Ding, Changxing and Xu, Xiangmin},
  year = 2024,
  pages = {28191--28201},
  doi = {10.1109/CVPR52733.2024.02663}
}

@inproceedings{li2024Pixels,
  title = {From pixels to graphs: Open-vocabulary scene graph generation with vision-language models},
  shorttitle = {From Pixels to Graphs},
  booktitle = {Proceedings of the IEEE/CVF Conference on Computer Vision and Pattern Recognition (CVPR)},
  author = {Li, Rongjie and Zhang, Songyang and Lin, Dahua and Chen, Kai and He, Xuming},
  year = 2024,
  pages = {28076--28086},
  doi = {10.1109/CVPR52733.2024.02652}
}

@inproceedings{liu2020ConsNet,
  title = {ConsNet: learning consistency graph for zero-shot human-object interaction detection},
  shorttitle = {ConsNet},
  booktitle = {Proceedings of the ACM International Conference on Multimedia},
  author = {Liu, Ye and Yuan, Junsong and Chen, Chang Wen},
  year = 2020,
  pages = {4235--4243},
  doi = {10.1145/3394171.3413600}
}

@inproceedings{luo2024Discovering,
  title = {Discovering Syntactic Interaction Clues for Human-Object Interaction Detection},
  booktitle = {Proceedings of the IEEE/CVF Conference on Computer Vision and Pattern Recognition (CVPR)},
  author = {Luo, Jinguo and Ren, Weihong and Jiang, Weibo and Chen, Xi'ai and Wang, Qiang and Han, Zhi and Liu, Honghai},
  year = 2024,
  pages = {28212--28222},
  doi = {10.1109/CVPR52733.2024.02665}
}

@article{zhang2025survey,
  title={A survey of reinforcement learning for large reasoning models},
  author={Zhang, Kaiyan and Zuo, Yuxin and He, Bingxiang and Sun, Youbang and Liu, Runze and Jiang, Che and Fan, Yuchen and Tian, Kai and Jia, Guoli and Li, Pengfei and others},
  journal={arXiv preprint arXiv:2509.08827},
  year={2025}
}

@article{luo2025empirical,
  title={An empirical study of catastrophic forgetting in large language models during continual fine-tuning},
  author={Luo, Yun and Yang, Zhen and Meng, Fandong and Li, Yafu and Zhou, Jie and Zhang, Yue},
  journal={IEEE Transactions on Audio, Speech and Language Processing},
  year={2025},
  publisher={IEEE}
}

@article{ma2024FGAHOI,
  title = {FGAHOI: fine-grained anchors for human-object interaction detection},
  shorttitle = {Fgahoi},
  author = {Ma, Shuailei and Wang, Yuefeng and Wang, Shanze and Wei, Ying},
  year = 2024,
  journal = {IEEE Transactions on Pattern Analysis and Machine Intelligence},
  volume = {46},
  number = {4},
  pages = {2415--2429},
  doi = {10.1109/TPAMI.2023.3331738}
}

@inproceedings{ning2023HOICLIP,
  title = {HOICLIP: Efficient Knowledge Transfer for HOI Detection with Vision-Language Models},
  shorttitle = {Hoiclip},
  booktitle = {Proceedings of the IEEE/CVF Conference on Computer Vision and Pattern Recognition (CVPR)},
  author = {Ning, Shan and Qiu, Longtian and Liu, Yongfei and He, Xuming},
  year = 2023,
  pages = {23507--23517},
  doi = {10.1109/CVPR52729.2023.02251}
}

@inproceedings{park2023ViPLO,
  title = {ViPLO: Vision Transformer Based Pose-Conditioned Self-Loop Graph for Human-Object Interaction Detection},
  shorttitle = {ViPLO},
  booktitle = {Proceedings of the IEEE/CVF Conference on Computer Vision and Pattern Recognition (CVPR)},
  author = {Park, Jeeseung and Park, Jin-Woo and Lee, Jong-Seok},
  year = 2023,
  pages = {17152--17162},
  doi = {10.1109/CVPR52729.2023.01645}
}

@inproceedings{tamura2021QPIC,
  title = {QPIC: query-based pairwise human-object interaction detection with image-wide contextual information},
  shorttitle = {Qpic},
  booktitle = {Proceedings of the IEEE/CVF Conference on Computer Vision and Pattern Recognition (CVPR)},
  author = {Tamura, Masato and Ohashi, Hiroki and Yoshinaga, Tomoaki},
  year = 2021,
  pages = {10405--10414},
  doi = {10.1109/CVPR46437.2021.01027}
}

@inproceedings{wang2024Bilateral,
  title = {Bilateral adaptation for human-object interaction detection with occlusion-robustness},
  booktitle = {Proceedings of the IEEE/CVF Conference on Computer Vision and Pattern Recognition (CVPR)},
  author = {Wang, Guangzhi and Guo, Yangyang and Xu, Ziwei and Kankanhalli, Mohan},
  year = 2024,
  pages = {27970--27980},
  doi = {10.1109/CVPR52733.2024.02642}
}

@inproceedings{wu2023Endtoend,
  title={End-to-end zero-shot hoi detection via vision and language knowledge distillation},
  author={Wu, Mingrui and Gu, Jiaxin and Shen, Yunhang and Lin, Mingbao and Chen, Chao and Sun, Xiaoshuai},
  booktitle={Proceedings of the AAAI conference on artificial intelligence},
  volume={37},
  number={3},
  pages={2839--2846},
  year={2023}
}

@inproceedings{wan2024exploiting,
  title={Exploiting CLIP for zero-shot HOI detection requires knowledge distillation at multiple levels},
  author={Wan, Bo and Tuytelaars, Tinne},
  booktitle={Proceedings of the IEEE/CVF Winter Conference on Applications of Computer Vision},
  pages={1805--1815},
  year={2024}
}

@inproceedings{wu2024Exploring,
  title = {Exploring Pose-Aware Human-Object Interaction via Hybrid Learning},
  booktitle = {Proceedings of the IEEE/CVF Conference on Computer Vision and Pattern Recognition (CVPR)},
  author = {Wu, Eastman Z Y and Li, Yali and Wang, Yuan and Wang, Shengjin},
  year = 2024,
  pages = {17815--17825},
  doi = {10.1109/CVPR52733.2024.01687}
}

@inproceedings{guo2024unseen,
  title={Unseen no more: Unlocking the potential of clip for generative zero-shot hoi detection},
  author={Guo, Yixin and Liu, Yu and Li, Jianghao and Wang, Weimin and Jia, Qi},
  booktitle={Proceedings of the ACM International Conference on Multimedia},
  pages={1711--1720},
  year={2024}
}

@inproceedings{wu2024Openset,
  title={Toward open-set human object interaction detection},
  author={Wu, Mingrui and Liu, Yuqi and Ji, Jiayi and Sun, Xiaoshuai and Ji, Rongrong},
  booktitle={Proceedings of the AAAI Conference on Artificial Intelligence},
  volume={38},
  number={6},
  pages={6066--6073},
  year={2024}
}

@inproceedings{xie2023Category,
  title = {Category query learning for human-object interaction classification},
  booktitle = {Proceedings of the IEEE/CVF Conference on Computer Vision and Pattern Recognition (CVPR)},
  author = {Xie, Chi and Zeng, Fangao and Hu, Yue and Liang, Shuang and Wei, Yichen},
  year = 2023,
  pages = {15275--15284},
  doi = {10.1109/CVPR52729.2023.01466}
}

@inproceedings{xiong2024Modalitycollaborative,
  title = {Modality-collaborative test-time adaptation for action recognition},
  booktitle = {Proceedings of the IEEE/CVF Conference on Computer Vision and Pattern Recognition (CVPR)},
  author = {Xiong, Baochen and Yang, Xiaoshan and Song, Yaguang and Wang, Yaowei and Xu, Chang Sheng},
  year = 2024,
  pages = {26722--26731},
  doi = {10.1109/CVPR52733.2024.02524}
}

@InProceedings{hoi4bot,
  title = 	 {HOI4ABOT: Human-Object Interaction Anticipation for Human Intention Reading Collaborative roBOTs},
  author =       {Mascaro, Esteve Valls and Sliwowski, Daniel and Lee, Dongheui},
  booktitle = 	 {Proceedings of The Conference on Robot Learning},
  pages = 	 {1111--1130},
  year = 	 {2023},
  editor = 	 {Tan, Jie and Toussaint, Marc and Darvish, Kourosh},
  volume = 	 {229},
  series = 	 {Proceedings of Machine Learning Research},
  month = 	 {06--09 Nov},
  publisher =    {PMLR},
  url = 	 {https://proceedings.mlr.press/v229/mascaro23a.html},
}

@inproceedings{cao2023Remine,
  title = {Re-mine, learn and reason: exploring the cross-modal semantic correlations for language-guided HOI detection},
  shorttitle = {Re-mine, learn and reason},
  booktitle = {Proceedings of the IEEE/CVF International Conference on Computer Vision (ICCV)},
  author = {Cao, Yichao and Tang, Qingfei and Yang, Feng and Su, Xiu and You, Shan and Lu, Xiaobo and Xu, Chang},
  year = 2023,
  pages = {23435--23446},
  doi = {10.1109/ICCV51070.2023.02147}
}

@inproceedings{liao2022GENVLKT,
  title = {GEN-VLKT: simplify association and enhance interaction understanding for HOI detection},
  shorttitle = {Gen-vlkt},
  booktitle = {Proceedings of the IEEE/CVF Conference on Computer Vision and Pattern Recognition (CVPR)},
  author = {Liao, Yue and Zhang, Aixi and Lu, Miao and Wang, Yongliang and Li, Xiaobo and Liu, Si},
  year = 2022,
  pages = {20091--20100},
  doi = {10.1109/CVPR52688.2022.01949}
}

@article{ma2025surveyvisionlanguageactionmodelsembodied,
  title={A survey on vision-language-action models for embodied ai},
  author={Ma, Yueen and Song, Zixing and Zhuang, Yuzheng and Hao, Jianye and King, Irwin},
  journal={arXiv preprint arXiv:2405.14093},
  year={2024}
}

@article{xue2025Zeroshot,
  title = {Towards Zero-shot Human-Object Interaction  Detection via Vision-Language Integration},
  author = {Xue, Weiying and Liu, Qi and Xiong, Qiwei and Wang, Yuxiao and Wei, Zhenao and Xing, Xiaofen and Xu, Xiangmin},
  year = 2025,
  journal = {neural networks},
  eprint = {2403.07246},
  primaryclass = {cs},
  archiveprefix = {arXiv}
}

@article{yang2023Boosting,
  title={Boosting human-object interaction detection with text-to-image diffusion model},
  author={Yang, Jie and Li, Bingliang and Yang, Fengyu and Zeng, Ailing and Zhang, Lei and Zhang, Ruimao},
  journal={arXiv preprint arXiv:2305.12252},
  year={2023}
}

@inproceedings{yang2024Openworld,
  title = {Open-world human-object interaction detection via multi-modal prompts},
  booktitle = {Proceedings of the IEEE/CVF Conference on Computer Vision and Pattern Recognition (CVPR)},
  author = {Yang, Jie and Li, Bingliang and Zeng, Ailing and Zhang, Lei and Zhang, Ruimao},
  year = 2024,
  eprint = {2406.07221},
  primaryclass = {cs},
  doi = {10.48550/arXiv.2406.07221},
  archiveprefix = {arXiv}
}

@inproceedings{yuan2022RLIP,
  title = {RLIP: relational language-image pre-training for human-object interaction detection},
  booktitle = {Proceedings of the Advances in Neural Information Processing Systems (NeurIPS)},
  author = {Yuan, Hangjie and Jiang, Jianwen and Albanie, Samuel and Feng, Tao and Huang, Ziyuan and Ni, Dong and Tang, Mingqian},
  year = 2022
}

@inproceedings{yuan2023RLIPv2,
  title = {RLIPv2: Fast Scaling of Relational Language-Image Pre-training},
  shorttitle = {RLIPv2},
  booktitle = {Proceedings of the IEEE/CVF International Conference on Computer Vision (ICCV)},
  author = {Yuan, Hangjie and Zhang, Shiwei and Wang, Xiang and Albanie, Samuel and Pan, Yining and Feng, Tao and Jiang, Jianwen and Ni, Dong and Zhang, Yingya and Zhao, Deli},
  year = 2023,
  pages = {21592--21604},
  doi = {10.1109/ICCV51070.2023.01979}
}

@inproceedings{zhang2022Efficient,
  title = {Efficient two-stage detection of human-object interactions with a novel unary-pairwise transformer},
  booktitle = {Proceedings of the IEEE/CVF Conference on Computer Vision and Pattern Recognition (CVPR)},
  author = {Zhang, Frederic Z. and Campbell, Dylan and Gould, Stephen},
  year = 2022,
  pages = {20072--20080},
  doi = {10.1109/CVPR52688.2022.01947}
}

@inproceedings{zhang2023exploring,
  title={Exploring predicate visual context in detecting of human-object interactions},
  author={Zhang, Frederic Z and Yuan, Yuhui and Campbell, Dylan and Zhong, Zhuoyao and Gould, Stephen},
  booktitle={Proceedings of the IEEE/CVF International Conference on Computer Vision (ICCV)},
  pages={10411--10421},
  year={2023}
}

@inproceedings{zhang2024HiKERSGG,
  title = {HiKER-SGG: hierarchical knowledge enhanced robust scene graph generation},
  shorttitle = {HiKER-SGG},
  booktitle = {Proceedings of the IEEE/CVF Conference on Computer Vision and Pattern Recognition (CVPR)},
  author = {Zhang, Ce and Stepputtis, Simon and Campbell, Joseph and Sycara, Katia and Xie, Yaqi},
  year = 2024,
  pages = {28233--28243},
  doi = {10.1109/CVPR52733.2024.02667}
}

@inproceedings{zheng2023OpenCategory,
  title = {Open-Category Human-Object Interaction Pre-training via Language Modeling Framework},
  booktitle = {Proceedings of the IEEE/CVF Conference on Computer Vision and Pattern Recognition (CVPR)},
  author = {Zheng, Sipeng and Xu, Boshen and Jin, Qin},
  year = 2023,
  pages = {19392--19402},
  doi = {10.1109/CVPR52729.2023.01858}
}

@article{bai2025Qwen25VL,
  title = {Qwen2.5-VL technical report},
  author = {Bai, Shuai and Chen, Keqin and Liu, Xuejing and Wang, Jialin and Ge, Wenbin and Song, Sibo and Dang, Kai and Wang, Peng and Wang, Shijie and Tang, Jun and others},
  year = {2025},
  month = feb,
  journal={arXiv preprint arXiv:2502.13923},
  eprint = {2502.13923},
  primaryclass = {cs},
  publisher = {arXiv},
  doi = {10.48550/arXiv.2502.13923},
  archiveprefix = {arXiv},
  langid = {english}
}

@article{achiam2023gpt,
  title={Gpt-4 technical report},
  author={Achiam, Josh and Adler, Steven and Agarwal, Sandhini and Ahmad, Lama and Akkaya, Ilge and Aleman, Florencia Leoni and Almeida, Diogo and Altenschmidt, Janko and Altman, Sam and Anadkat, Shyamal and others},
  journal={arXiv preprint arXiv:2303.08774},
  year={2023}
}

@article{qwen2025qwen25technicalreport,
  title={Qwen2.5 Technical Report},
  author={Yang, An and Yang, Baosong and Zhang, Beichen and Hui, Binyuan and Zheng, Bo and Yu, Bowen and Li, Chengyuan and Liu, Dayiheng and Huang, Fei and Wei, Haoran and others},
  journal={arXiv preprint arXiv:2412.15115},
  year={2024}
}

@article{zhu2025internvl3,
  title={Internvl3: Exploring advanced training and test-time recipes for open-source multimodal models},
  author={Zhu, Jinguo and Wang, Weiyun and Chen, Zhe and Liu, Zhaoyang and Ye, Shenglong and Gu, Lixin and Tian, Hao and Duan, Yuchen and Su, Weijie and Shao, Jie and others},
  journal={arXiv preprint arXiv:2504.10479},
  year={2025}
}

@article{liu2023Visual,
  title={Visual instruction tuning},
  author={Liu, Haotian and Li, Chunyuan and Wu, Qingyang and Lee, Yong Jae},
  journal={Proceedings of the Advances in Neural Information Processing Systems (NeurIPS)},
  volume={36},
  pages={34892--34916},
  year={2023}
}

@article{ren2024Learning,
  title = {Learning self- and cross-triplet context clues for human-object interaction detection},
  author = {Ren, Weihong and Luo, Jinguo and Jiang, Weibo and Qu, Liangqiong and Han, Zhi and Tian, Jiandong and Liu, Honghai},
  year = 2024,
  journal = {IEEE Transactions on Circuits and Systems for Video Technology},
  volume = {34},
  number = {10},
  pages = {9760--9773},
  doi = {10.1109/TCSVT.2024.3402247}
}

@article{touvron2023llama,
  title={Llama: Open and efficient foundation language models},
  author={Touvron, Hugo and Lavril, Thibaut and Izacard, Gautier and Martinet, Xavier and Lachaux, Marie-Anne and Lacroix, Timoth{\'e}e and Rozi{\`e}re, Baptiste and Goyal, Naman and Hambro, Eric and Azhar, Faisal and others},
  journal={arXiv preprint arXiv:2302.13971},
  year={2023}
}

@article{vaswani2017Attention,
  title = {Attention is all you need},
  author = {Vaswani, Ashish and Shazeer, Noam and Parmar, Niki and Uszkoreit, Jakob and Jones, Llion and Gomez, Aidan N. and Kaiser, Lukasz and Polosukhin, Illia},
  year = {2017},
  month = aug,
  journal={arXiv preprint arXiv:1706.03762},
  eprint = {1706.03762},
  primaryclass = {cs},
  publisher = {arXiv},
  doi = {10.48550/arXiv.1706.03762},
  archiveprefix = {arXiv},
  langid = {english}
}

@article{zhang2023VideoLLaMA,
  title={Video-llama: An instruction-tuned audio-visual language model for video understanding},
  author={Zhang, Hang and Li, Xin and Bing, Lidong},
  journal={arXiv preprint arXiv:2306.02858},
  year={2023}
}

@article{zhang2021mining,
  title={Mining the benefits of two-stage and one-stage hoi detection},
  author={Zhang, Aixi and Liao, Yue and Liu, Si and Lu, Miao and Wang, Yongliang and Gao, Chen and Li, Xiaobo},
  journal={Proceedings of the Advances in Neural Information Processing Systems (NeurIPS)},
  volume={34},
  pages={17209--17220},
  year={2021}
}

@misc{hicodet,
      title={Learning to Detect Human-Object Interactions}, 
      author={Yu-Wei Chao and Yunfan Liu and Xieyang Liu and Huayi Zeng and Jia Deng},
      year={2018},
      eprint={1702.05448},
      archivePrefix={arXiv},
      primaryClass={cs.CV},
      url={https://arxiv.org/abs/1702.05448}, 
}

@inproceedings{zhan2024enhancinghoidetectioncontextual,
  title={Enhancing HOI Detection with Contextual Cues from Large Vision-Language Models},
  author={Zhan, Yu-Wei and Liu, Fan and Luo, Xin and Xu, Xin-Shun and Nie, Liqiang and Kankanhalli, Mohan},
  booktitle={Proceedings of the ACM International Conference on Multimedia},
  pages={8557--8566},
  year={2025}
}

@article{yang2025qwen3technicalreport,
  title={Qwen3 technical report},
  author={Yang, An and Li, Anfeng and Yang, Baosong and Zhang, Beichen and Hui, Binyuan and Zheng, Bo and Yu, Bowen and Gao, Chang and Huang, Chengen and Lv, Chenxu and others},
  journal={arXiv preprint arXiv:2505.09388},
  year={2025}
}
}


\end{document}